\documentclass[runningheads]{llncs}
\usepackage{graphicx}
\usepackage{amsmath,amssymb} 
\usepackage{color}
\usepackage{booktabs}
\usepackage{multirow}
\usepackage{floatrow}
\usepackage{makecell}
\usepackage{hhline}

\newfloatcommand{capbtabbox}{table}[][]
\begin{document}
\pagestyle{headings}
\mainmatter

\title{Peeking Behind Objects: Layered Depth Prediction from a Single Image} 

\titlerunning{Peeking Behind Objects: Layered Depth Prediction from a Single Image}
\authorrunning{Dhamo et al.}

\author{Helisa Dhamo\inst{1}, Keisuke Tateno\inst{1,2}, Iro Laina\inst{1}, \\ Nassir Navab\inst{1}, Federico Tombari\inst{1}}

\institute{Technical University of Munich, Germany \\
\email{\{dhamo, tateno, laina, navab, tombari\}@in.tum.de}\\
\and Canon Inc., Tokyo, Japan}

\maketitle

\begin{abstract}

While conventional depth estimation can infer the geometry of a scene from a single RGB image, it fails to estimate scene regions that are occluded by foreground objects. 
This limits the use of depth prediction in augmented and virtual reality applications, that aim at scene exploration by synthesizing the scene from a different vantage point, or at diminished reality. 
To address this issue, we shift the focus from conventional depth map prediction to the regression of a specific data representation called Layered Depth Image (LDI), 
which contains information about the occluded regions in the reference frame and can fill in occlusion gaps in case of small view changes. We propose a novel approach based on Convolutional Neural Networks (CNNs) to jointly predict depth maps and foreground separation masks used to condition Generative Adversarial Networks (GANs) for hallucinating plausible color and depths in the initially occluded areas. We demonstrate the effectiveness of our approach for novel scene view synthesis from a single image. 

\end{abstract}

\section{Introduction}

With the increasing availability of new hardware in the market of virtual reality and user-machine interaction, such as Head-Mounted Displays (HMD), various solutions have been explored to make traditional image-based content available for such devices. The advantage would be 3D immersive visualization of a huge amount of visual data which is either already available on web databases (Google, Flickr, Facebook) or commonly acquired by consumer cameras present on smartphones and tablets. 
One relevant and, recently, actively investigated direction is monocular depth prediction \cite{Eigen:2014:DMP:2969033.2969091,Eigen:2015:PDS:2919332.2919917,laina2016deeper,liu2015,li15,Liu14,Wang_2015_CVPR,xu}, that aims to regress the scene geometry from a single RGB frame, to provide 3D content that can be viewed via stereo HMDs. 

\begin{figure}[!t]
\centering
\includegraphics[width=0.95\linewidth]{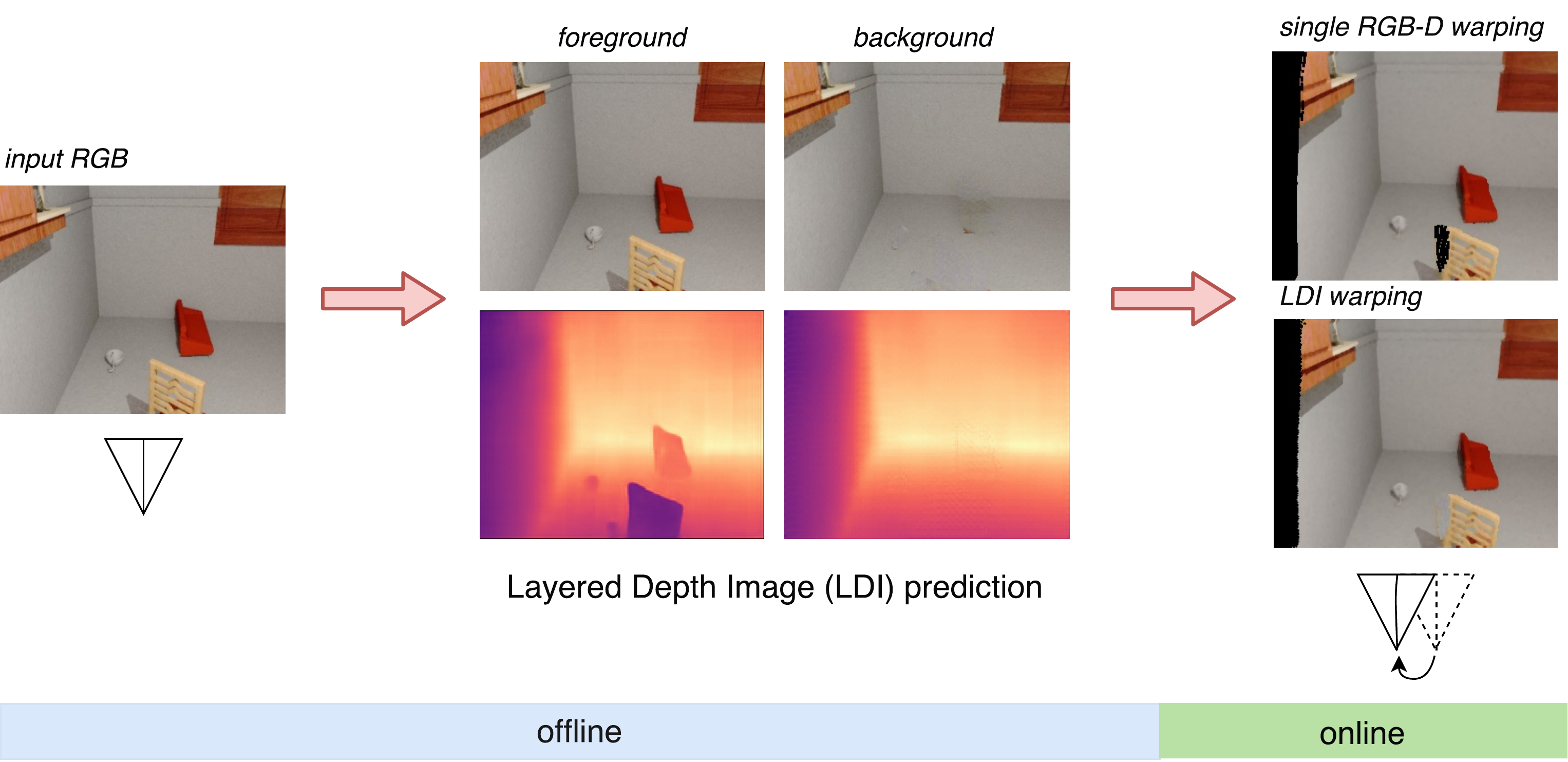}
\caption{ (\textit{Left}) Single RGB input in the original view. (\textit{Center}) Layered Depth Image prediction using the proposed approach. The \textit{foreground} layer consists of the original RGB image and monocular depth prediction; the \textit{background} RGB-D layer describes the same scene after foreground object diminishing. (\textit{Right}) One application of our method: viewpoint perturbation of the original image comparing two cases, simple warping of a single RGB-D and rendering from an LDI representation. The LDI generation is done offline, prior to the online simulation of the viewpoint changes.}
\label{fig:teaser}
\end{figure}

Nevertheless, the scene geometry obtained via depth prediction is relative to the specific viewpoint associated to the RGB frame. When viewed through an HMD, viewpoint changes induced by the user's head motion would reveal holes in the predicted geometry due to the occlusion caused by foreground objects (Fig. \ref{fig:teaser}, \textit{right}).
An alternative solution would be to predict a full 3D reconstruction of the scene from a single viewpoint. Nevertheless, this is a hard task to achieve due to the inherent ambiguity. Most related work is limited to reconstruction of objects only \cite{choy20163d,Fan_2017_CVPR,3dgan}, or requires depth as that acquired by an active 3D sensor \cite{FirmanCVPR2016,song2016ssc}. Also, the 3D volume of the scene is computation- and memory-intensive to process, which seems excessive for our goal, since for this type of applications we aim to solve the missing geometry only for small viewpoint changes. 
 
 An efficient compromise between 2.5D and 3D, would be an RGB-D structure that contains two layers, namely the first visible color and depth, and the next values along the same ray that are not part of the same object. This data representation was originally introduced by Shade et al. \cite{Shade:1998:LDI:280814.280882} as Layered Depth Image (LDI). While rendering from a target view, the RGB and depths from the second LDI layer are utilized to fill in the holes that become present after perturbing the viewpoint. Recently, Hedman et al. \cite{Casual3D2017} improve upon the original LDI structure to construct panoramic 3D photography from multiple views. However, in practical situations the only input source is a color image, for which neither tracking nor depth data is available. Hence, we propose a novel approach to generate an LDI for a similar application, but restricting the input to a single RGB, as illustrated in Fig. \ref{fig:teaser}. To deal with this added ambiguity, we make a \textit{simple scene assumption}, meaning that there is no more than one level of occlusion in every pixel of any view. In particular, an advantage of our approach for applications such as HMD-based virtual reality is \textit{efficiency}, as the LDI can be computed once (offline) and online warping is accomplished by directly processing the LDI data with low computational effort on the fly. This differs from the field of view synthesis, where regions occluded from the available vantage point require inpainting online for every new viewpoint. 

 To the best of our knowledge, we are the first to infer an LDI representation from a single RGB image. As a first step of our pipeline (Fig. \ref{fig:pipeline}) we learn a standard depth map and a foreground--background mask using a fully--convolutional network. Second, using the latter prediction, we remove the foreground in the RGB and depth images. This results in incomplete RGB-D background images, thus we use a GAN-based approach to inpaint the missing regions. Our contribution is not defined by specific deep learning tools within our framework. In fact, our method is agnostic to individual inpainting and depth prediction architectures. However, we introduce a \emph{pair discriminator} to encourage inter-domain consistency (here RGB-D). Besides view synthesis, the generated RGB-D with removed foreground, offers the potential for a range of AR applications, such as rendering new content in a diminished scene.  

\section{Related Work}
\label{sec:related}

To our knowledge, there is no research work on LDI prediction from a single RGB image. However, the work of this paper is built upon certain methods and data representations, which we want to visit in detail.

{\bf Layered Depth Images (LDI)} encompass a scene representation that contains multiple depth values along each ray line in a single camera view. It was first introduced by Shade et al. \cite{Shade:1998:LDI:280814.280882}, as an efficient image-based rendering method, emerged in times of limited computational power. When rendering in a perturbed view, some scene structures that are occluded in the original camera frame become visible -- known as disocclusion -- and therefore populate the target depth map with additional information. Applications include video view interpolation \cite{Zitnick:2004:HVV:1186562.1015766} and 3D photography \cite{Casual3D2017}. 
Recently, Hedman et al. \cite{Casual3D2017} present a 3D photo capturing from a set of hand--held camera images, that incorporates a cost term into a plane--sweep MVS, to penalize close depths. The different views are stitched together in a panorama LDI of two layers, which can be subject of geometry aware effects. Liu et al. \cite{liu2016layered} decompose a scene in depth layers (given an RGB-D input), with no color information being inferred in occluded regions.      

In contrast to all previous approaches, our application assumes a single input image, which makes the problem of obtaining an LDI more challenging.

{\bf Single-view depth estimation} One of the very first approaches for monocular depth estimation involved hand-engineered features and inference via Markov Random Fields \cite{saxena2006learning,saxena2009make3d}. Data-driven methods were later proposed that involved transferring and warping similar candidates from a database \cite{karsch2016depth}. Recently, the application of CNNs for depth map prediction (2.5D) from a single RGB image is widespread. In the pioneering work of Eigen et al. \cite{Eigen:2014:DMP:2969033.2969091}, a multi-scale CNN is proposed for coarse and refined maps, further extended to normals and semantics \cite{Eigen:2015:PDS:2919332.2919917}. Methods based on deeper networks further boost performance. Laina et al. \cite{laina2016deeper} introduce a ResNet--based \cite{DBLP:journals/corr/HeZRS15} fully convolutional network and a reverse Huber loss. Kendall and Yal \cite{kendall2017uncertainties} propose a Bayesian fully convolutional DenseNet \cite{huang2017densely} for capturing uncertainty. In contrast, regression forests with shallow CNNs as tree nodes are proposed in \cite{Roy2016MonocularDE}. Another family of methods \cite{li15,Liu14,xu} use Conditional Random Field regularizers to encourage geometrical consistency.

Besides 2.5D, CNNs have been also applied to 3D model prediction. Choy et al. \cite{choy20163d} map an RGB image to a 3D volume, using a 2D encoder and 3D decoder architecture, as well as a recurrent update component in case of multiple inputs. Fan et al. \cite{Fan_2017_CVPR} instead, predict a 3D point cloud from an RGB, by generating multiple 3D shapes. Wu et al. \cite{3dgan} (3D--VAE--GAN), builds a 3D model from the latent vector of an image. All these approaches output an object 3D model only, as opposed to our goal of representing whole scenes. Recently, Tulsiani et al. \cite{factored3dTulsiani17} introduce learning of a 3D scene representation that consists of a room layout and a set of object shapes. This method does not provide color information for the occluded background, as required in our proposed VR application. 

Our LDI prediction lies on 2.75D, between these two major branches of CNN-based 3D perception.  

{\bf View synthesis} The task of predicting the depth or 3D structure of a scene is further related to novel view synthesis and we later demonstrate our LDI prediction within this challenging application. View synthesis has been extensively addressed in prior work. A recent line of work on synthesizing new views directly minimizes a reconstruction loss between the reference and the target image in an end-to-end manner \cite{flynn2016deepstereo,zhou2016view,xie2016deep3d}. Besides that, depth prediction often arises as an implicit, intermediate representation in such frameworks when using view pairs as input/supervision \cite{xie2016deep3d,garg2016unsupervised,godard2017unsupervised,zhou2017unsupervised}.

{\bf CNN-based image inpainting} The goal of image inpainting (completion) is to fill in missing parts of an image. In the deep learning era, inpainting has been typically done with CNNs \cite{NIPS2015_5774,Yang_2017_CVPR}. 
More recently, Generative Adversarial Networks (GANs) \cite{NIPS2014_5423} have also been used for a variety of image prediction tasks, among which
inpainting \cite{Pathak_2016_CVPR,pix2pix2016,iizuka2017globally}, 
editing \cite{zhu2016generative} and 
3D shape filling \cite{wang2017shape}, 
due to their success in producing realistic looking samples.
In essence, a GAN consists of a generator $G$ and a discriminator $D$, trained with conflicting objectives. G aims to generate realistic data, whereas D distinguishes samples from the real data distribution. 
Conditional GANs \cite{mirza2014conditional,radford2015unsupervised} use additional priors, such as a lower resolution image \cite{DBLP:journals/corr/LedigTHCATTWS16}, or an incomplete image \cite{Pathak_2016_CVPR} to control the generation. Isola et al. \cite{pix2pix2016} propose a general--purpose GAN for a variety of image translation tasks, including inpainting. In this work, we extend the image completion task to RGB-D data, which to our knowledge has not been tackled before.

\section{Method}
\label{sec:method}

\begin{figure}[!t]
\centering
\includegraphics[width=0.8\linewidth]{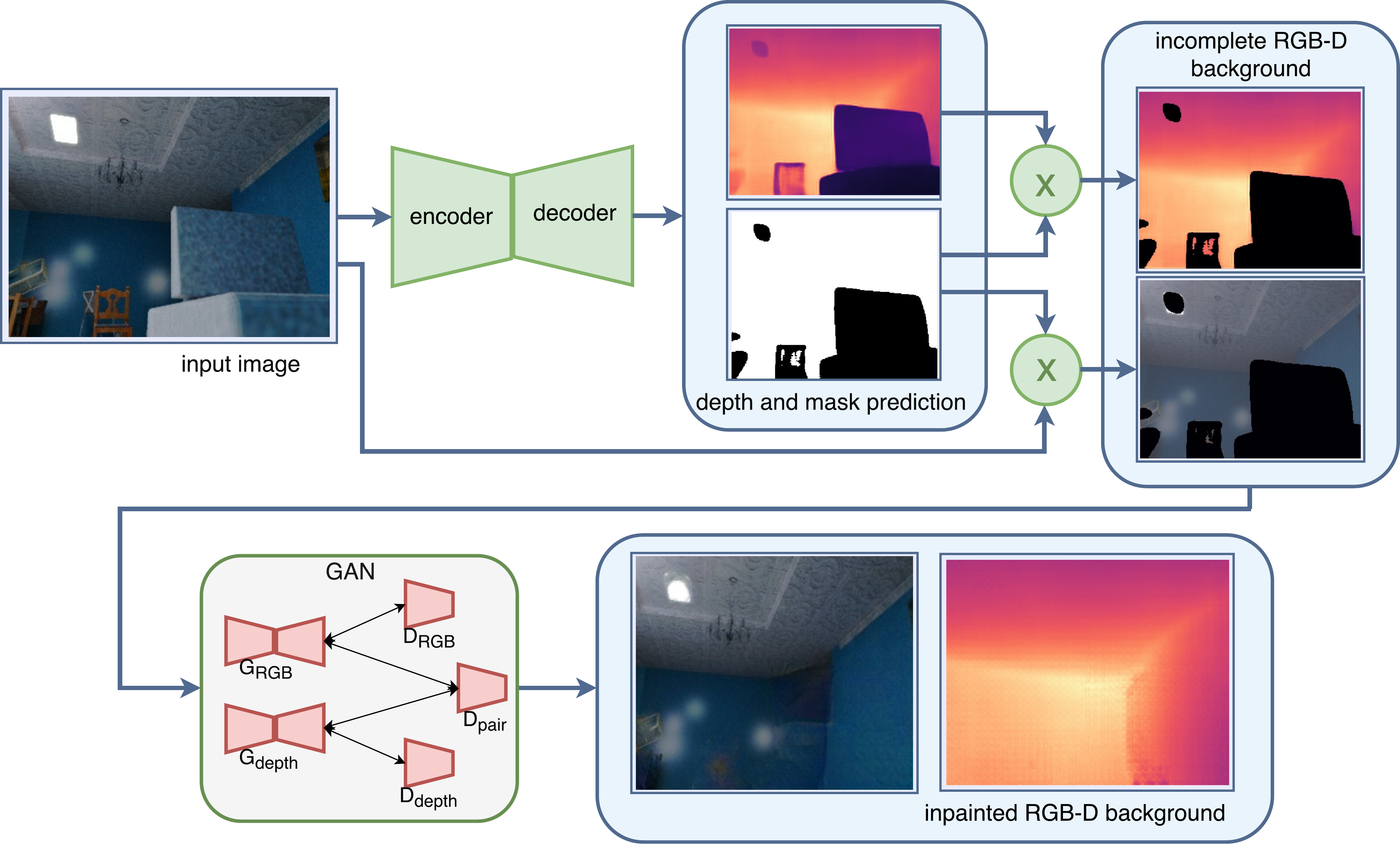}
\caption{General pipeline of our proposed method. (\textit{Top}) A single layer depth map and segmentation mask is predicted, using a fully--convolutional CNN. The segmentation mask is applied to the RGB input as well as the predicted depth, so that the respective foreground pixels are discarded. (\textit{Bottom}) The incomplete RGB-D information is inpainted using a GAN approach}
\label{fig:pipeline}
\end{figure}

Our goal is to learn a mapping from a single RGB image to an LDI, i.e. a two--layer RGB-D representation, as depicted in Fig.~\ref{fig:teaser}. In Section \ref{sec:dataset} we describe the acquisition of the data which will be used for the purpose of this work, as well as the accompanying assumptions. The proposed pipeline is presented in Fig.~\ref{fig:pipeline}. To break down the ambiguity of the prediction task, we suggest splitting the LDI prediction task into two subsequent steps. First, as explained in Section \ref{visible}, we train a network to jointly predict a conventional depth map and a foreground--background segmentation mask. Then, the mask is applied on both the RGB image and the predicted depth, so that only the visible background regions are fed to the next stage, while foreground regions are discarded. Finally, we perform GAN-based background completion conditioned on both the incomplete RGB as well as depth samples, as detailed in Section \ref{gan}.

\subsection{Dataset generation}
\label{sec:dataset}

Since learning to predict layered depth maps from a single image is a novel task, publicly available datasets containing LDI representations are quite limited for deep learning purposes. For instance, the authors of \cite{Casual3D2017}, currently provide a dataset of only 20 panoramic 3D photos \footnote{http://visual.cs.ucl.ac.uk/pubs/casual3d/datasets.html}, which have an LDI structure. Hence, we generate our own data based on existing datasets that provide RGB-D sequences with associated camera poses. 

We construct the LDI samples by projecting multiple views of the same scene onto a reference frame. This populates the color and depth representations of the reference frame with new depth values, which correspond to background regions that were not visible in the original depth map. To obtain a robust warping from a source frame to a target frame, the camera pose at each frame has to be known in advance. In our experiments, we used around $30000$ RGB-D frames with the associated camera poses. In addition, we build our dataset under the assumption of \textit{no self--occlusion}, i.e. an object is not occluding parts of itself, to avoid additional ambiguity in the prediction task. For this reason we also require object instance annotations, to make sure that at a specific coordinate, RGB-D values that correspond to the same object are not stored in both foreground and background layer. Given all these requirements, we generate our dataset based on SceneNet \cite{scenenet:ICCV2017} data, that consists of synthesized photorealistic environments.
 
The LDI extraction procedure is as follows. For each subset of $N=20$ consecutive RGB-D frames extracted from the same scene, we define the middle frame as reference, $f_{ref}$. Then, we warp every pixel $\boldsymbol{u}$ of the supportive frames $f_i$ into the reference view
\begin{equation}
\label{eq:warp}
\mathbf{u}_\mathrm{warped} = \pi \left(\mathbf{T}^\mathrm{ref}_{i}\pi^{-1} \left(\mathbf{u}_{i}\right)\right) \; ,
\end{equation}
where $\mathbf{T}^\mathrm{ref}_{i}$ is the transformation from  $f_i$ to $f_{ref}$, and $\pi()$ denotes perspective projection. We use $\mathbf{u}_\mathrm{warped}$ as the reference frame coordinates of the re-projected colors, instance annotations and depths. 
The instance annotations are particularly useful during warping to keep track of the scene entities responsible for occlusions.

Along with pixel coordinates, the perspective projection gives the new depth values, corresponding to metric distance from the reference camera plane. Note that SceneNet \cite{scenenet:ICCV2017} provide ray lengths instead of depths, therefore the following conversion is necessary
 \begin{equation}
 d(\mathbf{u}) = \frac{ r(\mathbf{u})}{||\mathbf{K}^{-1}\mathbf{\dot u}||_2}
 \end{equation}
 between the ray length $r$ and depth $d$, where $\mathbf{K}$ is the camera intrinsic matrix and $ \mathbf{\dot u} $ denotes a pixel in homogeneous coordinates. 
  
After computing the depth values for the occluded background regions of the reference view, we want to populate the background layer. The following validity conditions apply: 
  \begin{enumerate}
 	\item  The candidate depth value is larger compared to the respective foreground pixel at that coordinate.
 	 \begin{equation}
    d_\mathrm{warped}(\mathbf{u}_\mathrm{warped}) > d_\mathrm{ref}^\mathrm{FG}(\mathbf{u}_\mathrm{warped})
    \end{equation}
 	\item  The candidate background pixel and the respective original foreground pixel do not share the same object $id$. Here, we rely on the assumption of \emph{no self--occlusion}.
 	\begin{equation}
    id_\mathrm{warped}(\mathbf{u}_\mathrm{warped}) \neq id_\mathrm{ref}^\mathrm{FG}(\mathbf{u}_\mathrm{warped})
    \end{equation}
 	\item The candidate pixel is a potential background layer pixel, only if the object instance that contains it does not occlude other objects at any point. In such a way, we make sure that the \emph{simple scene assumption} is fulfilled, meaning that an object instance can not be part of both layers.
 \end{enumerate}
 
\begin{figure}[!t]
    \centering
    \includegraphics[width=0.98\linewidth]{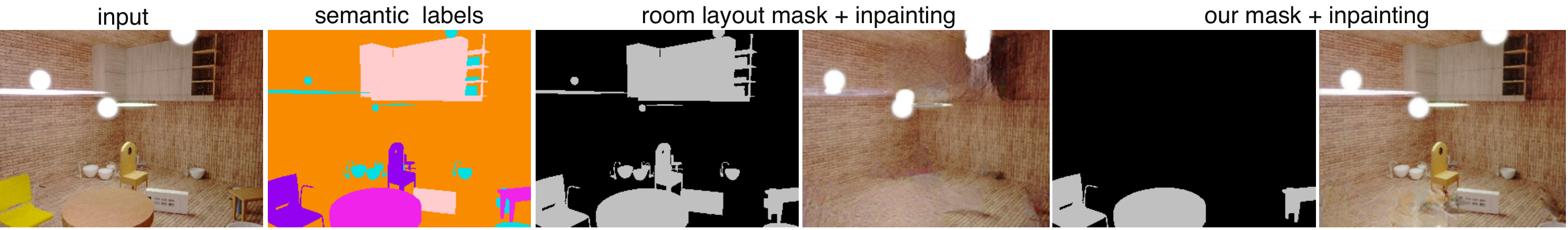}
    \caption{Illustration of semantic room layout separation vs.~our foreground segmentation method.}
    \label{fig:room_layout}
    \vspace{-0.5em}
\end{figure}

Finally, we store the warped RGB-D frame with the smallest depth among all valid candidates in the background layer. Correspondingly, we extract a foreground--background segmentation mask, utilizing the accumulated information on object instances. Namely, a pixel in the image is considered as background if its reference instance label is not found in the list of occluding instances, and foreground otherwise. As a result, as described in Section \ref{gan}, the background inpainting will not be prone to undesired foreground context. We use the original \emph{train/valid} split of SceneNet, to generate our training and test data. 
It is worth noting that the result of our method differs from a simple room layout separation, which would instead give a 3D ``box'' representation of the scene. In particular some of the object instances should still be considered as part of the background context, given an application that involves \emph{small} viewpoint perturbations. For instance, in Fig. \ref{fig:room_layout} we expect to see more parts of the occluded chair instead of plain floor, while exploring around the table in front of it. 

Importantly, we observed that this LDI reconstruction approach does not work on real datasets like ScanNet \cite{dai2017scannet}, in which instance segmentation is not perfectly accurate. This leads to almost every pixel in the image being classified as foreground, as long as there is a single background pixel in the instance annotation maps, that is wrongly classified as foreground. Therefore, while for the current ground truth generation we rely on synthetic data, our method does also generalize to real-world data as shown later through experimental evaluation. 

\subsection{Joint depth map and segmentation mask prediction}
\label{visible}

The first stage of the proposed pipeline consists in standard depth map estimation as well as foreground--background segmentation from a single RGB image. 

For this task, a wealth of CNN architectures for depth prediction and/or semantic segmentation exists in literature and could be employed \cite{Eigen:2014:DMP:2969033.2969091,Eigen:2015:PDS:2919332.2919917,laina2016deeper,liu2015,li15,xu}. 
In our work, we employ the fully convolutional ResNet-50 architecture proposed by Laina et al. \cite{laina2016deeper} given its competitive performance in depth estimation and the publicly available implementation\footnote{https://github.com/iro-cp/FCRN-DepthPrediction}. We modify the original network architecture by adding one more up--projection layer, so that the output depth preserves the input resolution.
By converting the ray length images originally provided by SceneNet to metric distances, we get depth maps corresponding to the \textit{visible} structures in the LDI representation of every frame, that is the \textit{first layer} composing the LDI. 
We additionally aim to learn a segmentation map that would later allow to mask out the foreground entities in the current reference view, thus creating holes for the upcoming background completion. 
We recognize that this is not a deterministic task, especially in complex scenes, since there might be multiple valid separation options. However, we train the same model for segmentation, considering the binary masks we generated in Section \ref{sec:dataset} as ground truth. This can be interpreted as \emph{implicit} learning of an adaptive threshold for the separation, dependent on the distance from the camera, occlusion, as well as the structure continuity. For example, in Fig. \ref{fig:pipeline}, wall regions that are closer compared to the brown chair, are still predicted as background, since they are smoothly connected with the rest of the wall and do not cause occlusion.   

We train our fully convolutional network on the depth prediction and foreground segmentation tasks, both separately and jointly. The latter shows superior performance for both tasks, which is intuitively also justified by the underlying relation between them. Our final model configuration allows the two tasks to share all network weights, except from the last layer. The loss function for the depth component is the reverse Huber $\mathcal{B}(d)$, following \cite{laina2016deeper}, whereas for the foreground segmentation $L_2$--norm $\mathcal{L}_2(s)$ outperformed the standard cross entropy loss. We train jointly by combining the depth loss and segmentation loss with equal weight:
\begin{equation}
\mathcal{L}_\mathrm{total}= \mathcal{B}(d) + \mathcal{L}_2(s).
\end{equation}
After obtaining the continuous segmentation predictions, we apply a threshold of $0.45$ to distinguish between background and foreground. The threshold favors classification as foreground to prevent the subsequent background completion from getting confused by undesired foreground influence, while losing part of the background context does not have a negative effect on completion performance. Moreover, we observe that applying dilation on the resulting masks, further removes undesired foreground information, usually located around the object borders. Concretely, in our tests, we used cross--shaped dilation structure with a size of $5$ pixels. 
 
\subsection{GAN-based RGB-D inpainting}
\label{gan}

Depth prediction \emph{behind} occlusion brings additional ambiguity compared to conventional depth map prediction, since it implies inferring distance where the respective color context is not visible from a single view. Hence, learning depths for the unknown is rather a hallucinatory process, that involves creating plausible missing context conditioned on the visible regions as a prior. 

In this work, we exploit the inpainting potential with a GAN-based approach. 
We start from a state-of-the-art model as a baseline and explore plausible ways of boosting the accuracy while extending to the RGB-D case. We use a similar architecture as in Isola et al.~\cite{pix2pix2016} for the generator. Our discriminators adopt the C64-C128-C256-C512 architecture as proposed in \cite{pix2pix2016}, where C denotes a Convolution--BatchNorm--ReLU block followed by the number of filters. The base loss in our inpainting GANs is
\begin{equation}
\label{eq:gan_loss}
\mathcal{L}_\mathrm{inpaint} = \mathbb{E}_{x,y}[log D(x,y)] + 
\mathbb{E}_{x}[log(1-D(x, G(x)))] 
\end{equation}

\noindent where $x$ represents incomplete input and $y$ the corresponding full image. In this formulation, G optimizes the following objective
\begin{equation}
\hat G = \min\limits_G \max\limits_D \mathcal{L}_\mathrm{inpaint} + \lambda _{\mathcal{L}_1} \mathcal{L}_1(y - G(x)).
\end{equation}

\noindent The $\mathcal{L}_1$ loss accounts for similarity between generated and ground truth samples. 

Intuitively, one would argue that a joint learning between color and depth could potentially enhance the consistency between them. However, the semantic diversity between the two, encourages separate learning. We explore different architectures, to investigate the adverse responses of these two motivations.

First, we train a GAN for \textit{combined RGB-D completion}. In this approach, the color and depth counterparts share all the generator and discriminator weights, leading to the lowest number of parameters. The objective function $\mathcal{L}_\mathrm{RGB-D}$ has the form of Eq. \ref{eq:gan_loss}, with $x = x_{RGB-D}$ and $y = y_{RGB-D}$. In addition, we explore \textit{separate RGB and depth completion}, without any shared parameters between the respective RGB and depth inpainting, i.e. $\mathcal{L}_\mathrm{RGB}$ and $\mathcal{L}_\mathrm{depth}$ are updated independently from each other. Finally, we introduce an additional GAN model, built upon the latter and further enhanced with a multi-modal \emph{pair discriminator} $D_\mathrm{pair}$, whose role is to encourage inter-domain consistency between RGB and depth. We refer to it as \textit{separate RGB and depth completion with pairing}. $D_\mathrm{pair}$ takes an RGB-D input, either from the ground truth or the generator and distinguishes real RGB and depth \emph{correspondences}. The respective generators $G_\mathrm{depth}$ and $G_{c}$, receive feedback from their individual discriminators as well as from the combined domain, thus optimizing an additional term
\begin{equation}
\mathcal{L}_\mathrm{pair} = \mathbb{E}_{x,y}[log D_\mathrm{pair}(y_{c},y_{d})] + 
\mathbb{E}_{x}[log(1-D_\mathrm{pair}(G_{c}(x_{c}), G_{d}(x_{d})))],
\end{equation}

\noindent where $x_c$ and $y_c$ refer to the color image, and $x_d$ and $y_d$ to depth. The final $\hat {G}_{c}$ and $\hat {G}_{d}$ objectives then become
\begin{equation}
\hat G_\mathrm{c} = \min\limits_{G_{c}} \max\limits_{D_{RGB}} + \lambda _\mathrm{pair} \min\limits_{G_{c}} \max\limits_{D_{pair}} \mathcal{L}_\mathrm{pair} + \lambda _{\mathcal{L}_1} \mathcal{L}_1(y_{c}, G_{c}(x_{c}))
\end{equation}


\begin{equation}
\hat G_\mathrm{d} = \min\limits_{G_{d}} \max\limits_{D_{d}} + \lambda _\mathrm{pair} \min\limits_{G_{d}} \max\limits_{D_{pair}} \mathcal{L}_\mathrm{pair} + \lambda _{\mathcal{L}_1} \mathcal{L}_1(y_{d}, G_{d}(x_{d}))
\end{equation}

\noindent with $\lambda_\mathrm{pair}=0.5$ and $\lambda _{L_1}=100$. In all our GAN models at hand, after normalizing the input color and depth images separately to $[-1,1]$, we set to $-2$ the values of inpainting interest. The GAN implicitly learns to identify and invalidate image regions assigned to those values. 

\section{Experiments}

In this section, we present the experiments we conducted to evaluate the performance of our proposed method on SceneNet \cite{scenenet:ICCV2017} (synthetic) and NYU depth v2 \cite{nyu2012} (real) datasets. For SceneNet, we create our ground truth background data, as described in Section \ref{sec:dataset}, therefore we perform both qualitative and quantitative measurements. Since this is not the case for NYU, we present qualitative results only, on the view synthesis application (Section \ref{sec:view_synthesis}), so that the reader can perceive the effect of the perturbed views in a real world context.  

\subsection{Conventional depth and background mask}
\label{sec:depth_error}

We train the joint depth map and foreground segmentation prediction task using a subset of around 30,000 RGB-D samples from the \emph{train} partition of the SceneNet dataset. Further, we test on around 500 images, from the \emph{valid} partition. For the depth prediction task we obtain a relative error of $0.184$. The segmentation result is evaluated using the intersection over union metric (IoU), which is 0.71 for foreground and 0.93 for background pixels.

\begin{figure}[!t]
\centering
\includegraphics[width=0.98\linewidth]{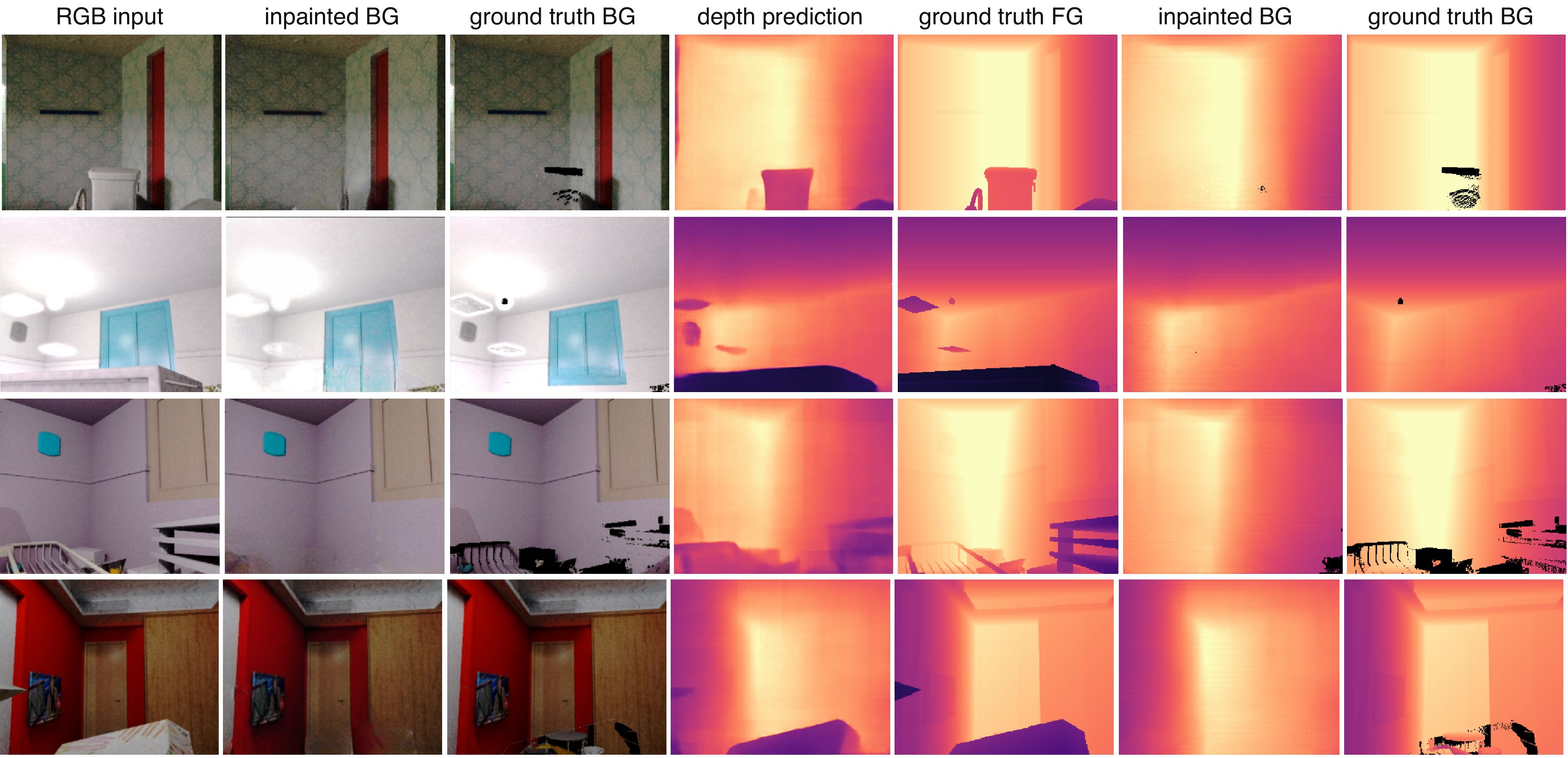}
\caption{Examples of foreground removal and background inpainting, for RGB and depth, on the SceneNet \cite{scenenet:ICCV2017} dataset, with accompanying ground truth. Black indicates invalid pixels on the RGB and depth maps}
\label{fig:fore_dim}
\end{figure}
\subsection{Ablation study on foreground diminishing}

Table \ref{table:results_pred} shows the quantitative results on the predicted background RGB and depth maps of our proposed pipeline. In absence of previous similar work, we formulate this evaluation process as an ablation study, to consider our main investigated inpainting approaches. We compute the relative error (rel) and root mean square error (rms) for depth, and structural similarity index (SSIM) for RGB which gives an accuracy measure alongside the rms error. We measure the errors in two different cases, one is for the entire image while the other considers only the background information that was predicted by the inpainting stage. Please note that here we compare the quality of the depth completion against ground truth data, despite it being conditioned on the depth prediction results from the CNN. Thus, all results are products of our pipeline that operates on just a single image. In addition, Table \ref{table:results_gt} illustrates the same evaluations when directly inpainting incomplete ground truth depth maps, where ground truth segmentation masks are also used to separate the foreground.

The ablation study shows that the \textit{combined RGB-D inpainting} approach performs the poorest in terms of depth prediction. In contrast, the accuracy of the RGB inpainting has a slight advantage compared to the other methods. One could argue, that this is due to the dominance of the RGB counterpart as textures and features from the color domain could have a negative impact on depth. 
Moreover, we believe that along with the incomplete RGB channels, the learning benefits from the incomplete depth information, as a better orientation for object separation. On the other end, the \textit{separate RGB and depth} approaches, with and without pairing have a clear advantage in terms of depth inpainting, particularly on the newly added background region. 
Additionally, our pairing loss term brings an improvement in the depth inpainting task, particularly when learning from predicted depths. Since the \textit{pair discriminator} encourages consistency between the complete RGB and depth maps, the depth map generator gets optimized to drive its output towards a plausible ground truth depth. 

\setlength{\tabcolsep}{4pt}
\begin{table}[!t]
\begin{center}
\caption{Analysis of our GAN versions, on the SceneNet dataset \cite{scenenet:ICCV2017}. Inpainting applied on \textit{predicted FG-BG masks and depths}, from the CNN model from \cite{laina2016deeper} }
\label{table:results_pred}
\begin{tabular}{l|c|cc|cc}
\hline
\multirow{3}{*}{Method} & \multirow{3}{*}{Image area} & \multicolumn{2}{c}{RGB metrics} & \multicolumn{2}{|c}{Depth error}\\
\cline{3-6}
 & & ssim & rms & rel & rms \\
 & & (higher) & (lower)  & (lower)  & (lower) \\
\hline
\multirow{2}{*}{Combined RGB-D}  & \textit{whole} & 0.935 & \textbf{17.59} & 0.156 & 0.574\\
& \textit{inpainted} & -- & \textbf{54.46} & 0.197 & 0.704 \\
\hline
\multirow{2}{*}{Separate RGB and depth} & \textit{whole} & 0.940 & 18.61 & 0.151 & 0.555\\
& \textit{inpainted} & -- & 58.51 & 0.175 & 0.633 \\
\hline
\multirow{2}{*}{Separate RGB and depth, paired}  & \textit{whole} & \textbf{0.942} & 18.04 & \textbf{0.149} & \textbf{0.549}\\
& \textit{inpainted} & -- & 56.73 & \textbf{0.172} & \textbf{0.622} \\

\hline
\end{tabular}
\end{center}
\end{table}
\setlength{\tabcolsep}{1.4pt}

\setlength{\tabcolsep}{4pt}
\begin{table}[!t]

\begin{center}
\caption{Analysis of our GAN versions, on the SceneNet dataset \cite{scenenet:ICCV2017}. Inpainting module applied on \textit{ground truth FG-BG masks and depths}}
\label{table:results_gt}
\begin{tabular}{l|c|cc|cc}
\hline
\multirow{3}{*}{Method} & \multirow{3}{*}{Image area} & \multicolumn{2}{c}{RGB metrics} & \multicolumn{2}{|c}{Depth error}\\
\cline{3-6}
 & & ssim & rms & rel & rms \\
 & & (higher) & (lower)  & (lower)  & (lower) \\

\hline

\multirow{2}{*}{Combined RGB-D}  & \textit{whole} & 0.891 & \textbf{19.45} & 0.044 & 0.189\\
& \textit{inpainted} & -- & \textbf{51.32} & 0.090 & 0.349 \\
\hline
\multirow{2}{*}{Separate RGB and depth} & \textit{whole} & 0.900 & 20.09 & 0.018 & 0.100\\
& \textit{inpainted} & -- & 53.98 & \textbf{0.041} & \textbf{0.193} \\
\hline
\multirow{2}{*}{Separate RGB and depth, paired}  & \textit{whole} & \textbf{0.903} & 19.76 & \textbf{0.017} & \textbf{0.095}\\
& \textit{inpainted} & -- & 52.70 & \textbf{0.041} & 0.197 \\

\hline
\end{tabular}
\end{center}
\end{table}
\setlength{\tabcolsep}{1.4pt}

Comparing the results of Table \ref{table:results_pred} and \ref{table:results_gt}, one can observe that the depth inpainting accuracy goes hand in hand with the accuracy of the predicted depth images. Notably, the relative error between the ground truth background and the completed map (Table \ref{table:results_gt}) is in negligible range, whereas the relative errors in Table \ref{table:results_pred} are not far from the inherent depth prediction error, reported in Section \ref{sec:depth_error}. Conventional depth maps, usually have a good absolute scale, but bring high error values around the object borders, due to the lack of sharpness in edges. This can explain why the relative errors from inpainting results, with removed foreground, are even lower than those of original depth map prediction. Interestingly, this means that LDI prediction from an RGB image, is an additional asset that can be easily incorporated into a regular depth prediction, without affecting the accuracy. In particular, recent advances in CNN depth prediction lead to more accurate LDIs, with no major change in the proposed method.

Fig. \ref{fig:fore_dim} illustrates background inpainting examples, using the CNN predicted depths and masks. As can be seen, the background ground truth is not always available, since during the warping procedure of Section \ref{sec:dataset}, some regions have been not visible from any of the available viewpoints. Nevertheless, GAN inpainting covers the whole image. From the second example, we observe that although the inpainted image has hallucinated a door as opposed to the ground truth window, the result is equally plausible, considering the uncertainty. 

\subsection{View synthesis evaluation}
\label{sec:view_synthesis}

To demonstrate the visual effect of a support background layer, we simulate a view perturbation scenario. In this experiment, the reference view corresponds to the original input RGB image. For the sake of simplicity, we define the reference camera pose to be in the origin of the world coordinates, meaning identity matrix rotation $\mathbf{R}_{ref}$ and zero translation $\mathbf{t}_{ref}$. We are interested in seeing the same content, from a slightly perturbed view angle. Concretely, we modify $t_x$ or $t_y$ in the reference translation vector $\mathbf{t}_{ref}$ to obtain a target pose with horizontal or vertical shift respectively. Next, we utilize Eq. \ref{eq:warp} to warp the color and depth images of both LDI layers, from the source view to the target views. The foreground layer is warped first, followed by a dilation and erosion to fill small holes. Afterwards, the corresponding background layer is warped, substituting the pixel values that are still void after the morphological operations. Note that the respective background layer in this experiment is the inpainted output of the learned depths, which exposes the rendering task to additional inaccuracy. 

\begin{figure}
\centering
\includegraphics[width=\linewidth]{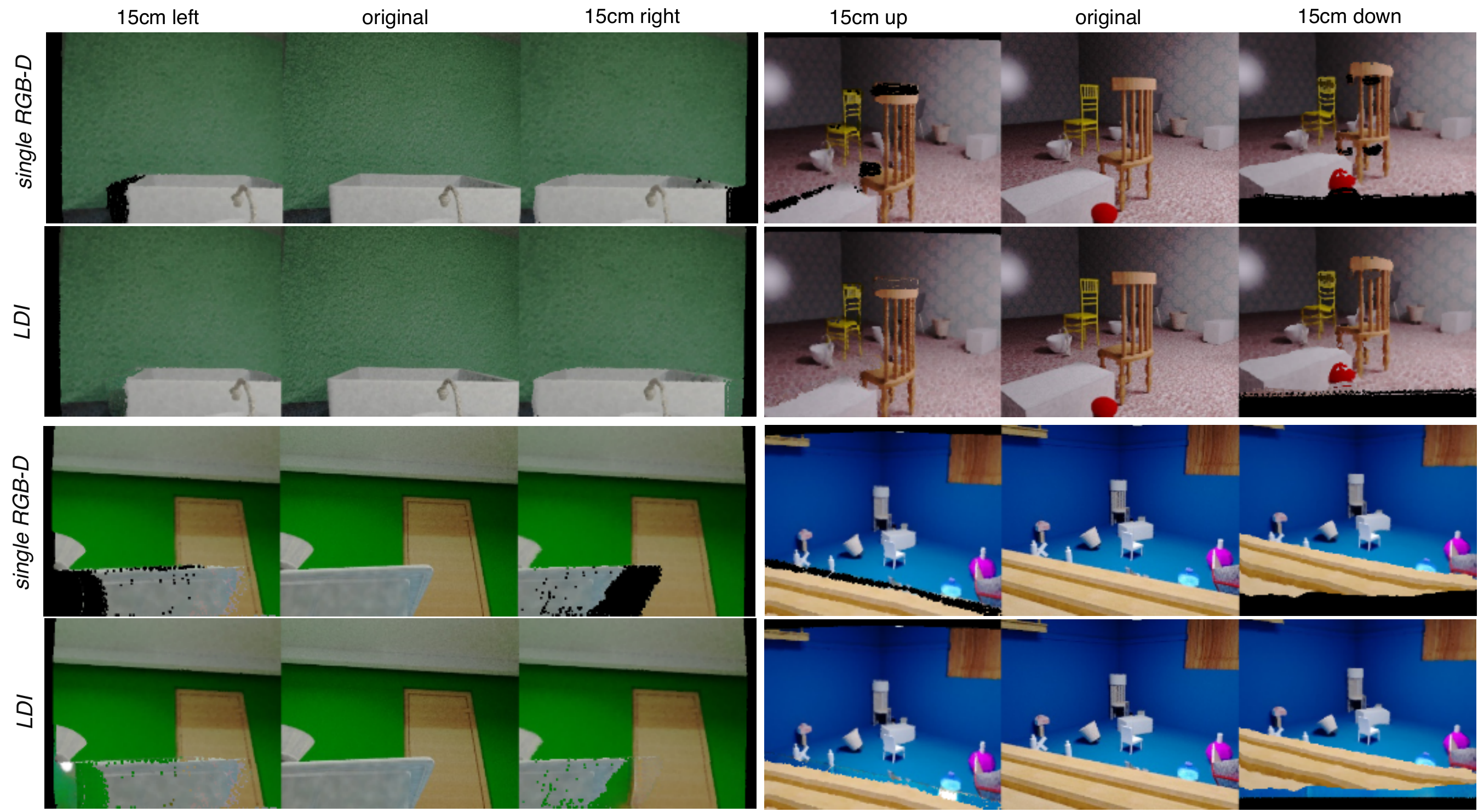}
\caption{Results on SceneNet \cite{scenenet:ICCV2017}, LDIs rendered on perturbed views. (\textit{Upper row}) for each scene presents the warping of a simple RGB-D layer. (\textit{Lower row}) shows the warping of the two layer RGB-D, obtained with our proposed pipeline. }
\label{fig:rendering_scenenet}
\end{figure}

\begin{figure}
\centering
\includegraphics[width=0.98\linewidth]{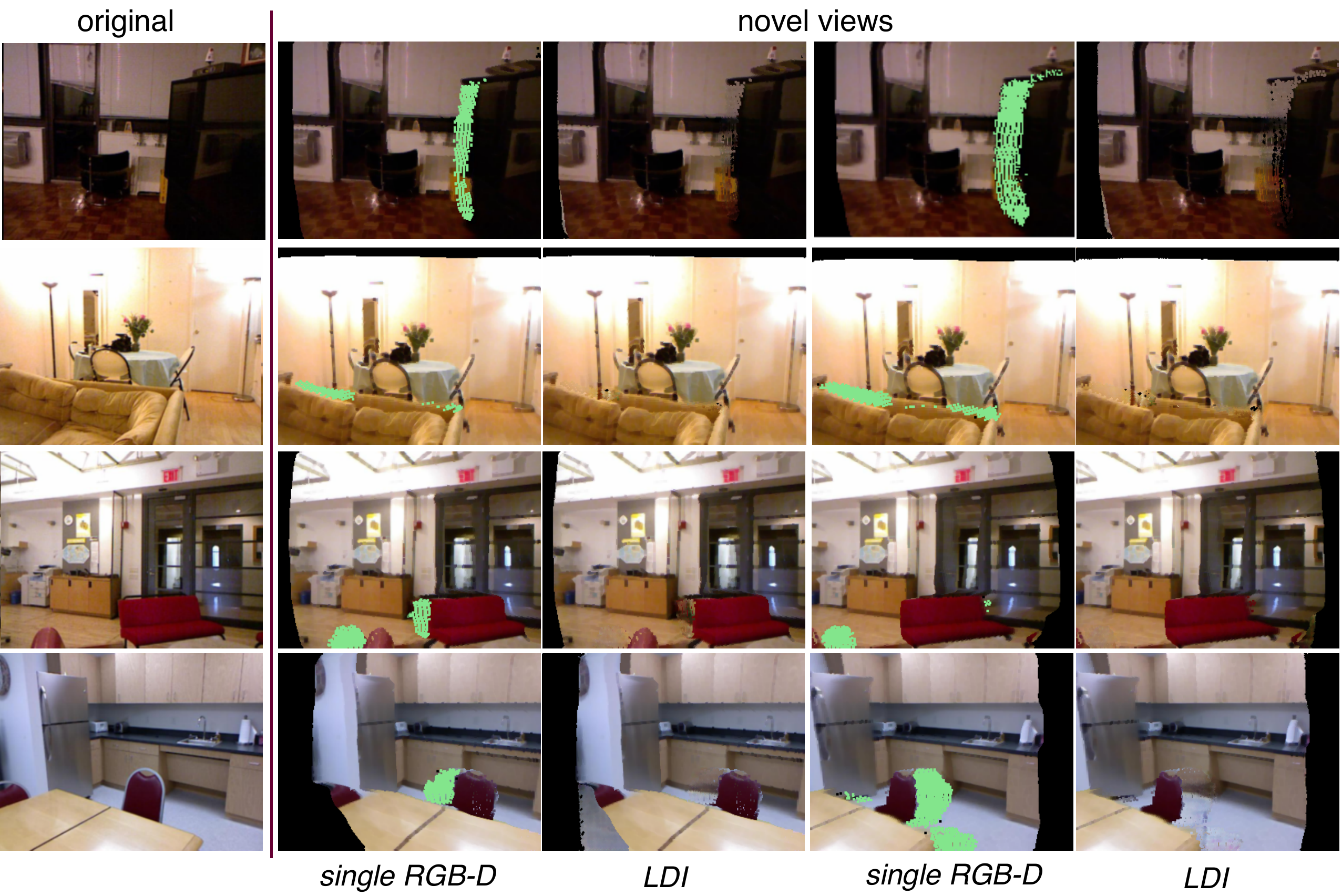}
\caption{Results on NYU dataset \cite{nyu2012}, LDIs rendered on perturbed views. (\emph{Left}) Input RGB image. (\emph{Right})  Two novel views obtained after rendering from a single RGB-D and our LDI representation. }
\label{fig:rendering_nyu}
\end{figure}

For every test frame, we simulate different perturbation levels, in all four directions (top, bottom, left and right).  The rendering results on the SceneNet dataset \cite{scenenet:ICCV2017} are shown in Fig. \ref{fig:rendering_scenenet}, whereas the equivalent results on NYU \cite{nyu2012} are to be seen in Fig. \ref{fig:rendering_nyu}. One can observe the added value in completing the target view with additional pixels, that are occluded in the reference view. The completion task adapts to the background context, even when edges or relatively complex structures and patterns are present in the occluded background.

In almost every example, one can observe missing pixels around the image border, on the side towards which the viewpoint change was made. This is due to the fact that the current method does not perform inpainting outside the image borders. In some cases, background color is available outside the view, and it appears in the rendered image even though it is out of place. This is to be witnessed for instance in the last example of Fig. \ref{fig:rendering_nyu}. Notably, dealing with inpainting outside the borders is not an easy task, since one has to hallucinate foreground information alongside with the background counterpart. Please refer to the paper supplement for more qualitative evaluations (images, video).   

\begin{figure}[h!]
\hspace{-3mm}
\begin{floatrow}
\capbtabbox[4.8cm]{{
        \begin{tabular}{l c @{\hskip 0.2in} c}
            \toprule
            Method & MAE & SSIM \\
            \midrule
            AF \cite{zhou2016view} & 0.200 & 0.537 \\
            AF \cite{zhou2016view} (FCRN)  & 0.185  & 0.534 \\
            Ours  & \textbf{0.147} & \textbf{0.617} \\
            \bottomrule
            \caption{Quantitative comparison with view synthesis methods.} 
            \label{table:results_vs}
        \end{tabular}
}}{}
    \ffigbox[\Xhsize]{
        \includegraphics[width = 7.2cm]{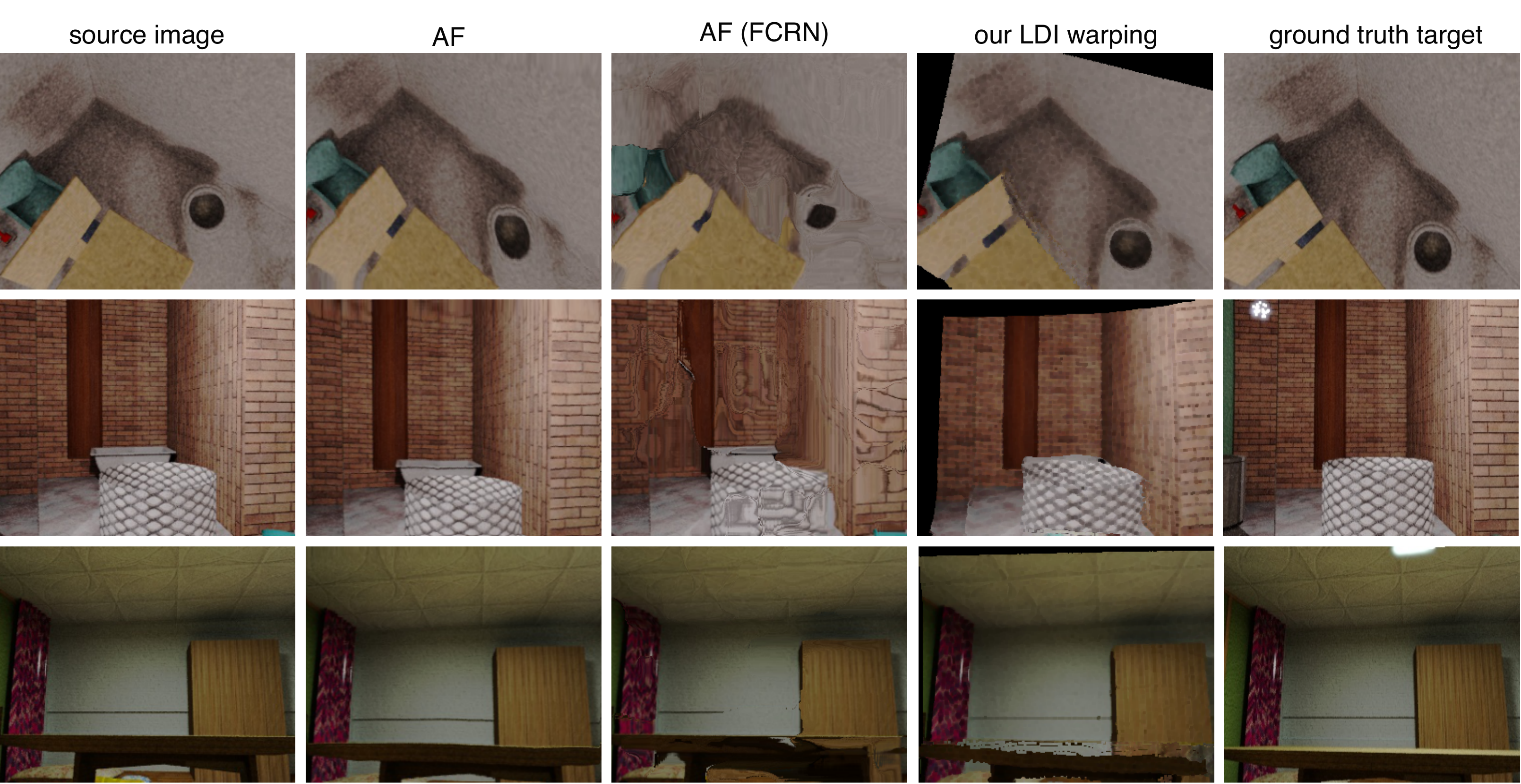}}{
        \caption{Qualitative comparison of novel view synthesis.}%
        \label{fig:vs_example}}
\end{floatrow}
\end{figure}

We performed comparative evaluations to state-of-the-art view synthesis work. To the best of our knowledge, the most comparable method is Appearance Flow (AF) \cite{zhou2016view} as it considers a single input image and an arbitrary viewpoint transformation. Apart from that, we note that the majority of the view synthesis literature explores multi-view scenarios  \cite{flynn2016deepstereo,xie2016deep3d,garg2016unsupervised,godard2017unsupervised,zhou2017unsupervised} or exploits scene geometry \cite{garg2016unsupervised,godard2017unsupervised,zhou2017unsupervised} (e.g.~emerging depth via stereo) and is thus limited by the learned geometry (e.g.~stereo baseline) for view generation.  
We trained their original model on SceneNet using the same partitions as for our own method. For fairness, we also trained AF based on a fully convolutional model (same as in our method \cite{laina2016deeper})\footnote{In their public repository, Zhou et al. \cite{zhou2016view} report better performance when switching to fully convolutional networks.}. We report the comparisons in Table \ref{table:results_vs} and show that the application of LDI to view synthesis outperforms AF \cite{zhou2016view} in two different metrics, Mean Absolute Error (MAE) and SSIM. Moreover, from Fig. \ref{fig:vs_example} one can see that our method preserves object shapes and aligns the image better with the target view.

\section{Conclusion}

We proposed a method to regress a layered depth map from a single RGB image. We illustrated how such maps can be regressed via a pipeline built over a CNN and a GAN. In addition we also demonstrate how the additional information included in a layered depth map can be useful for an enhanced user experience of the 3D content of a scene with respect to depth prediction, e.g. by  improving view synthesis under occlusion. Importantly, the quality of the LDI prediction goes along with the accuracy of the individual components of our pipeline, such as CNN-based depth prediction and GAN-based inpainting. 
Future work aims at learning to regress depth representations that support more than one level of occlusion, thus capable to overcome the \emph{simple scene assumption}. 

\bibliographystyle{splncs}
\bibliography{egbib}
\end{document}